\pdfoutput=1
\documentclass[11pt]{article}

\usepackage[final]{acl}

\usepackage{times}
\usepackage{latexsym}
\usepackage{multirow}
\usepackage{rotating}
\usepackage[T1]{fontenc}
\usepackage{amsmath}
\usepackage{amssymb}
\usepackage{booktabs} 
\usepackage{caption} 
\usepackage[utf8]{inputenc}

\usepackage{microtype}

\usepackage{inconsolata}

\usepackage{graphicx}

\usepackage{subcaption}

\title{Can you SPLICE it together? A Human Curated Benchmark for Probing Visual Reasoning in VLMs}

\author{
  \textbf{Mohamad Ballout}$^*$,
  \textbf{Okajevo Wilfred}$^*$,
  \textbf{Seyedalireza Yaghoubi},
\\  
  \textbf{Nohayr Muhammad Abdelmoneim},
  \textbf{Julius Mayer},
  \textbf{Elia Bruni}
\\
\\
  Institute of Cognitive Science, Osnabrück University, Osnabrück, Germany
\\
  \small{
    \href{mailto:email@domain}{mohamad.ballout@uos.de}
  }
}

\begin{document}
\maketitle
\def\thefootnote{*}\footnotetext{These authors contributed equally to this work}

\begin{abstract}

%
%Visual reasoning in multimodal large language models (MLLMs) remains significantly underexplored compared to reasoning in %large language models (LLMs). To address this gap, we introduce SPLICE, a human-curated benchmark designed to test multiple %aspects of reasoning in MLLMs by asking models to arrange event based cut clips from complete videos. SPLICE is derived from %the COIN instructional video dataset, consisting of 3,381 human-filtered videos segmented into distinct events, resulting in %11,423 clips. We evaluate both human participants and state-of-the-art models under two modalities: (1) vision-only (using %either raw videos or frames, depending on model capabilities) and (2) human-annotated summaries describing each clip.Our %results show that models struggle with visual reasoning, performing poorly when relying solely on vision. While human %performance remains consistent across both modalities, models exhibit a significant reliance on textual annotations to improve %their accuracy. However, since these annotations are human-crafted, this boost does not reflect genuine visual understanding. %Despite this added information, models still lag behind human performance.
%

% Visual reasoning in vision-language models (VLMs) remains significantly underexplored compared to reasoning in large language models (LLMs). 

In this work, we introduce SPLICE, a human-curated benchmark derived from the COIN instructional video dataset, designed to probe event-based reasoning across multiple dimensions: temporal, causal, spatial, contextual, and general knowledge. SPLICE includes 3,381 human-filtered videos spanning 12 categories and 180 sub-categories, such as sports, engineering, and housework. These videos are segmented into a total of 11,423 event clips. We evaluate both human participants and state-of-the-art vision-language models (VLMs) on the task of rearranging these clips into coherent event sequences to assess visual reasoning capabilities. Results reveal a significant gap: VLMs struggle to match human performance. While human-annotated textual descriptions improve model accuracy, they do not affect human performance, suggesting that models rely more on language priors than on visual understanding. Even with annotations, VLMs fall short of human-level reasoning, underscoring persistent challenges in visual reasoning. A deeper analysis across sub-categories shows that VLMs perform relatively better on videos where temporal and causal reasoning are dominant, compared to those where contextual and spatial reasoning are dominant. They also perform better on everyday tasks than on specialized ones.

% While this input improves VLM performance, it does not affect human performance, indicating that the text information is redundant for humans. However, the fact that the model's performance improves with textual input highlights a weakness in its inherent visual reasoning ability. 

\end{abstract}

\section{Introduction}

Transformer-based models  \citep{vaswani2017attention} initially focused on pre-training with language data alone \citep{radford2018improving,devlin2019bert,raffel2020exploring}. They later evolved to multi-modal pre-training with the introduction of patch-based training \citep{dosovitskiy2020image}. Since then, vision large language models (VLMs) have advanced rapidly, increasingly matching or even surpassing human performance across various domains, including coding, mathematics, scientific knowledge, and reasoning. For example, benchmarks like ARC-AGI \citep{chollet2019measure}, where models scored 0\% in 2019, now report state-of-the-art models achieving scores between 33\% and 55.5\% \citep{chollet2024arc}. While this remarkable progress in reasoning capabilities is essential for enhancing the utility of current AI systems, our understanding of how well these models reason about purely visual sequences remains incomplete. Unlike textual reasoning tasks, where progress is well-documented, the field lacks benchmarks that rigorously evaluate visual reasoning without heavy reliance on language priors.

% developing benchmarks that rigorously test the reasoning abilities of MLLMs has become both increasingly important and more challenging

\begin{figure}[t]
  \includegraphics[width=\columnwidth]{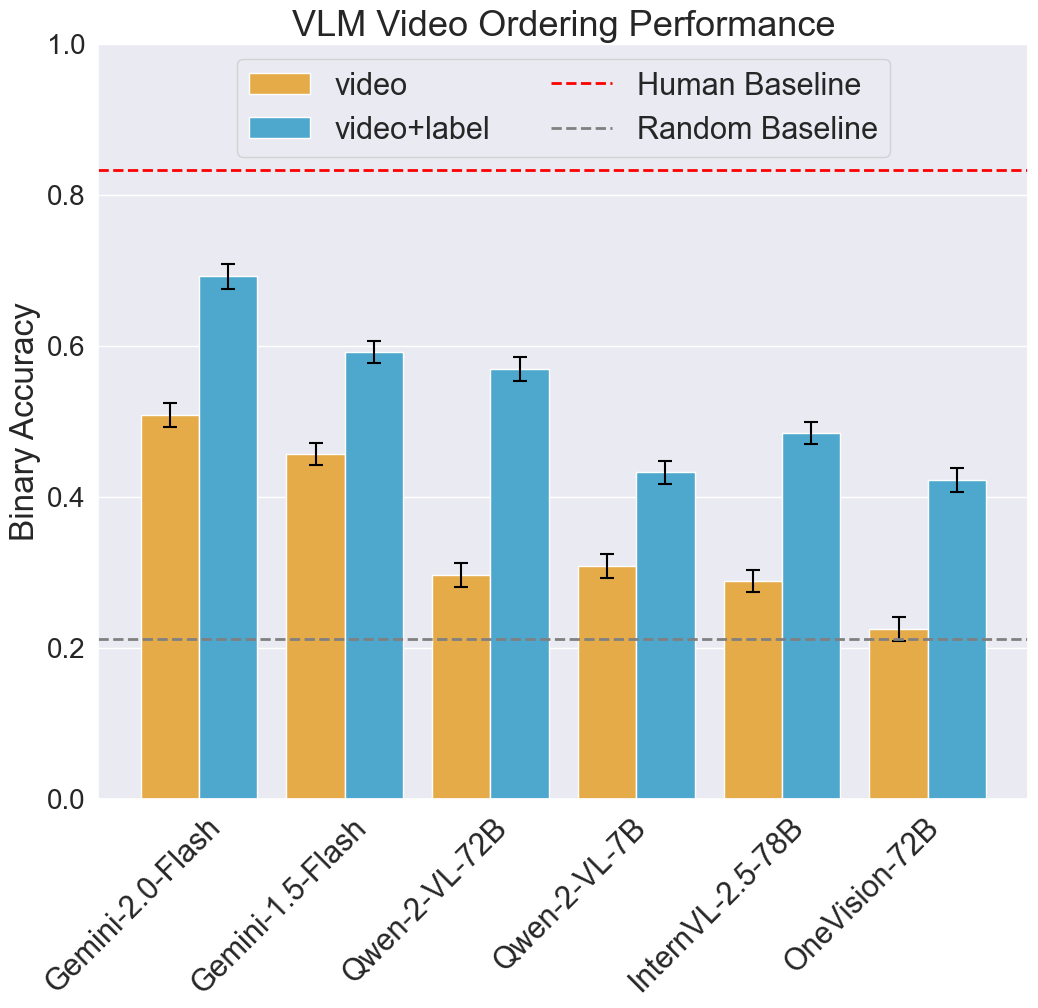}
  \caption{Video clip ordering accuracy of VLMs across 12 video categories (3,381 videos) in two modalities, compared to the human and random baseline.
}
  \label{fig:results}
\end{figure}

Alongside advancements in performance, VLMs have become significantly more efficient, enabling them to process long videos. For instance, open-source models like Qwen2-VL \citep{wang2024qwen2} can understand videos exceeding 20 minutes in length and handle multiple input videos simultaneously. However, despite these capabilities, current benchmarks do not sufficiently probe models' ability to infer event sequences from purely visual cues.

Leveraging this new capability, we propose an intuitive yet challenging benchmark, SPLICE\footnote{ \url{https://huggingface.co/datasets/prokajevo/splice-benchmark}. } (\textbf{S}equential \textbf{P}rocessing for \textbf{L}earning and \textbf{I}nference in \textbf{C}hronological \textbf{E}vents), where the task is to order shuffled clips, cut from original videos, based on the events depicted in them. SPLICE fills this gap by leveraging a dataset where the correct ordering of clips requires multiple types of reasoning, such as causal, temporal, spatial, contextual, and general knowledge reasoning.

Unlike previous datasets that rely on automatic or unsupervised segmentations, SPLICE is constructed through a rigorous human curation process. We adapt the COIN instructional video dataset \citep{tang2019coin}, which was originally created for video understanding and event localization, by extracting 3,600 videos spanning 180 tasks across 12 domains (e.g., vehicles, gadgets, cooking). In the "vehicles" domain, for instance, tasks may include changing tires, lights, or fuses. COIN provides event-based annotations, which we use to segment each video into distinct clips before shuffling them. This repurposing necessitates careful filtering and validation to ensure that only meaningful and well-defined tasks remain. Consequently, SPLICE eliminates ambiguous or trivial sequences, making it a stronger probe of true visual reasoning.

This event-based structure aims to prevent models from relying solely on the first and last frames or other shortcuts, instead requiring deeper reasoning. For instance, a Karate practice video may be divided into three clips: an opening salutation, practicing movements, and a closing salutation. Since the salutation clips are visually identical but occur at different points, the model must rely on other cues such as breathing, sweat levels, spatial positioning, or background actions to determine their order.

In this work, we compare the performance of multiple state-of-the-art models that support multi-video input, including Qwen2-VL \citep{wang2024qwen2}, Gemini-Flash \citep{gemini20flash}, InternVL2.5 \citep{chen2024expanding}, and LlavaOnevision \citep{li2024llava}, across three different input settings: videos only, text only, and videos+text. Additionally, we provide human performance benchmarks and compare them with the performance of these VLMs in both the video-only and video+text settings. Our results indicate that VLMs fall significantly behind humans, particularly in the vision-only setting, where there is a substantial performance gap.

The main contributions of this paper can be summarized as follows:

\begin{itemize}
\item We introduce  a simple and yet challenging, human-curated benchmark designed to test a model’s ability to reconstruct event sequences from shuffled video clips. The dataset consists of 3,381 human-validated videos, each segmented into multiple clips that must be ordered correctly.

\item We show that state-of-the-art models struggle with this task, achieving only 23\% to 51\% accuracy, while humans consistently score around 85\%.

\item We perform a reasoning-type analysis and show that VLMs perform better when the dominant type of reasoning is causal or temporal, as opposed to contextual or spatial.

\end{itemize}

\section{Related Work}

As this work evaluates VLMs on various aspects of reasoning, we provide a comprehensive review of reasoning evaluations for VLMs, along with tasks similar to our approach, where videos or images were shuffled and reordered.

\subsection{Reasoning in VLMs}

While reasoning in large language models has been extensively explored \citep{huang-chang-2023-towards, plaat2024reasoning, xu2025largereasoningmodelssurvey}, covering aspects such as temporal reasoning \citep{chu2023timebench, tang2024ltlbench, li2023unlocking, maruthi2022temporal}, causal reasoning \citep{zhang2023understanding, hobbhahn2022investigating}, general knowledge reasoning \citep{zhang2024chain, wu-etal-2024-reasoning}, and spatial reasoning \citep{li2024advancing, hu2024chain}, vision language models (VLMs) are considered relatively underexplored \citep{wang2024exploring}. This is partly due to their novelty, higher computational demands, and the complexity of evaluating their performance. Nevertheless, several studies have investigated different aspects of reasoning in VLMs \citep{wu2024mind, li2024multimodal, ko2023large, zhangmultimodal}.

%\begin{figure*}[t]
%  \centering
%  \includegraphics[width=\textwidth]{pic-acl-2.pdf} 
%  \caption{An example of a set of clips that the models need to order correctly. The figures show the first and last frames of each clip. The clips are segmented based on events, reducing the reliance on shortcuts. Clip durations vary based on the event, with gaps where moments not relevant to the main steps are omitted. In this video, models must infer that the oranges are cut, juiced, filtered, and then served. Clips order: c, a, d, b. }
%  \label{fig:sample}
%\end{figure*}

Several benchmarks have also been developed to evaluate VLMs on different reasoning aspects, such as intuitive physics \citep{jassim2023grasp, weihs2022benchmarking}, mathematics \citep{gupta2024polymath, chen2021geoqa}, spatial reasoning \cite{mayer2025ivispar}, and science in general \citep{bubeck2023sparks, nori2023capabilities}. With the increasing capabilities of VLMs and claims of achieving AGI, more challenging benchmarks—such as MMbench \citep{liu2025mmbench}, MMMU \citep{yue2024mmmu}, M4U \citep{wang2024m4u}, and ARC-AGI \citep{chollet2024arc}—have been introduced to assess the general capabilities of these models and compare them to human performance. In this context, we present our benchmark, which is human-baselined, to challenge the visual reasoning abilities of VLMs using diverse real-world instructional videos that require multiple aspects of reasoning to solve.

\subsection{Ordering Videos}

With the new capability of models to take multiple videos as input, most earlier work has focused on reordering videos \citep{misra2016shuffle, lee2017unsupervised, xu2019self, sharma2020deep, hu2021contrast, wang2023moviepuzzle} or images \citep{sevilla2021only, kim2023learning} based on their extracted embeddings. 

Additional works \citep{yang2025made, li2020hero, fernando2015modeling} in the literature use image or video ordering for pre-training or training models to enhance their temporal reasoning abilities. Other studies \citep{wei2018learning, zhou2015temporal} also employ video ordering, but with a primary focus on temporal reasoning. While our presented dataset could be used as a pre-training task, our main goal is to benchmark state-of-the-art vision-language models against human performance on event-based video ordering. This task serves as a proxy to evaluate not only temporal reasoning but also other types of reasoning, including causal, contextual, spatial, and general knowledge.

A closely related line of work \citep{xu2019self, hu2021contrast} treats clip reordering as a self-supervised task by extracting embeddings from uniformly sampled clips. In contrast, we input raw, non-overlapping clips directly into models, leveraging their ability to process multiple clips simultaneously. We also adopt an event-driven clip extraction strategy, where clip lengths vary based on event duration. This approach introduces reasoning challenges beyond temporal understanding, as discussed in Section \ref{depthreasoning}.

The approach in \citet{sharma2020deep} is similar to \citet{xu2019self} but incorporates additional modalities such as audio, text, and visual features. In our work, we also leverage the multi-modal capabilities of the models, employing two different settings: one using vision only, and another combining vision and text.

\section{The SPLICE Benchmark}

\subsection{Step-Annotated Video Dataset}
Our video ordering benchmark, SPLICE, is built on a subset of the COIN dataset \citep{tang2019coin}, a hierarchically organized collection of 11,827 instructional videos covering 180 tasks across 12 everyday domains\footnote{Domains include: Nursing \& Caring, Vehicles, Leisure \& Performance, Gadgets, Electric Appliances, Household Items, Science \& Craft, Plants \& Fruits, Snacks \& Drinks, Dishes, Sports, and Housework.}. Originally designed for event localization and video understanding, COIN provides detailed, human-annotated step-by-step segments with precise timestamps.

Each video in the COIN dataset is divided into multiple steps, with distinct actions occurring within specific time intervals. These steps were annotated by humans, specifying the exact timestamps (from time \(tx\) to \(ty\)) in the original video during which each action takes place.
A sample of a sequence of video is shown in Figure \ref{fig:sample}.

Although multiple videos cover the same task (e.g., preparing food), the dataset maintains diversity through variations in execution, step grouping (e.g., chopping and adding vegetables in one step), and real-world factors like camera angles, lighting, and backgrounds. 
This diversity and realism make the COIN dataset an excellent foundation for evaluating video ordering tasks, as it provides a challenging testbed for models to reason about temporal sequences and action segmentation in realistic instructional settings.

\subsection{From COIN to SPLICE: Data Preparation}

Due to the complexity and high cost of human cleaning and ordering, we randomly selected a 3,600-video subset from COIN before dividing them into smaller clips for ordering. This subset spans all domains and tasks, with each video originally containing up to 7 clips, segmented based on the original annotations.

The decision to limit the number of clips to 7 was motivated by several factors. First, computational memory constraints in the models make handling longer sequences impractical. Second, longer sequences are more challenging and time-intensive for humans to order accurately. Finally, many high-step videos in the COIN dataset involve repetitive loop actions (e.g., prolonged mixing), where it becomes infeasible to determine an order from shuffled clips, as there are no discernible changes after the action has been completed.

\subsection{Human Ordering Protocol}
Our dataset includes two modalities: videos-only and videos combined with text. The audio modality was excluded because not all models support audio alongside videos. We have 3,600 uncut videos in total, which effectively becomes 7,200 ordering tasks when considering both modalities. The videos were divided into eight sets of 900: four sets for the video-only modality and four sets for the video + text modality. Four annotators, all authors in this paper, were grouped into two teams, each consisting of a PhD student and a Master's student in cognitive science. Each team handled 1,800 videos across both modalities (900 per modality) without access to the original video order, ensuring that they relied solely on the provided clips for sequencing. The annotators used the following cross-checking procedure: 
\begin{itemize} 
\item \textbf{Annotator A:} orders 900 videos in the video-only modality and another 900 in the video + text modality, completing 1,800 ordering tasks in total.
\item \textbf{Annotator B:} orders the same 900 videos as Annotator A, but flips the modality for each set. Specifically, the 900 videos Annotator A ordered in video-only are ordered by Annotator B in video + text, and the 900 videos Annotator A ordered in video + text are ordered by Annotator B in video-only. 
\end{itemize}

Each set of 900 videos is ordered by both annotators in both modalities to enable cross-checking and ensure consistency. Videos are either ordered or marked as \textit{inconclusive}. Annotators provide their best guess regardless, and a video is labeled inconclusive only if both annotators independently agree. If one annotator provides an order and the other marks it inconclusive, the latter’s best guess is used as the final sequence.

We opted against crowd-sourcing to ensure transparency and consistency, as the publicly available dataset and metadata could compromise the reliability of crowd-sourced annotations.

\subsection{Criteria for Excluding Videos from Dataset}

Annotators were instructed to mark a video as \textit{inconclusive} only under the following circumstances:
\begin{enumerate}
\item \textbf{Repeated Instructions:} Identical actions performed separately (e.g., two different people demonstrating fire extinguisher use).
\item \textbf{Ambiguous Continuous Actions:} Actions spanning multiple clips without clear temporal cues for ordering (e.g., continuous CPR compressions).
\item \textbf{Unrelated Actions:} Different actions shown without contextual links to determine their sequence (e.g., cutting different vegetables without showing intermediate states).
\end{enumerate}
This systematic filtering ensured a high-quality dataset suitable for evaluating model reasoning. The final dataset comprises 3,381 videos, available in video-only and video+text modalities. Table \ref{tab:1} shows the video distribution by clip count; additional statistics about the videos are shown in Table \ref{tab:segment_distribution} in Appendix \ref{sec:vid-stats}.

% get the duration stats as well as catgories

\section{Types of Reasoning}
\label{resonsing}
Cutting, shuffling, and reordering clips may seem like a simple task requiring basic temporal reasoning. However, when applied to event-based instructional videos, it demands a richer set of reasoning skills. Below, we outline key reasoning types, with examples from our benchmark:

\textbf{Contextual Reasoning:}  Understanding the context of actions to predict what follows. E.g., when changing a printer cartridge, opening the door is typically followed by replacing the cartridge and then closing the door.

\textbf{Spatial Reasoning:} Interpreting spatial relationships and orientations. E.g., aligning a car to park or tracking an athlete’s movement across pole vault stages.

\textbf{General knowledge Reasoning:} Applying everyday knowledge to infer plausible actions. E.g., drying follows washing potatoes, or avoiding a hot plate after oven use.

\textbf{Temporal Reasoning:} Grasping the correct sequence of events. E.g., assembling a sofa by attaching legs before placing the mattress.

\textbf{Causal Reasoning:} Recognizing cause-effect links. E.g., mixing flour and water forms dough; cutting a bike chain results in it falling loose.

\begin{table}
  \centering
  \begin{tabular}{ccc}
    \hline
    \textbf{\# of Clips} & \textbf{\# of Videos}  & \textbf{Average Duration (s)} \\
    \hline
    2 &  1020 & 46.83         \\
    3 &  1026  & 53.31      \\
    4 &  734  & 62.41      \\
    5 &  333  & 72.67       \\
    6 &  172  & 67.86          \\
    7 &  96  & 73.50        
      \\\hline
      
  \end{tabular}
\caption{Distribution of videos by the number of clips, with a total of 3,381 videos segmented into 11,423 clips. The average duration per video is reported.}
\label{tab:1}
\end{table}

\section{Evaluation}
\subsection{Metrics}
As outlined earlier, each video in the dataset is segmented into clips based on the original COIN dataset's step localization. These clips are then randomly shuffled and renamed as \( C = \{c_1, c_2, \dots, c_n\} \).  Models were tested on different modalities, receiving  either video input, text annotations input  (short textual descriptions of events in the video) or a combination of both. The model's task is to predict the correct permutation of the clip order. Let \( \mathbf{y} = [y_1, y_2, \dots, y_n] \) denote the ground-truth sequence of clip indices, and \( \mathbf{\hat{y}} = [\hat{y}_1, \hat{y}_2, \dots, \hat{y}_n] \) represent the predicted sequence. For \( n \) clips, there exist \( n! \) possible permutations. We evaluate performance using two main metrics. 

\textbf{Binary Accuracy}. The prediction is correct only if the entire sequence matches the ground truth:
\[
\text{Binary Accuracy} = 
\begin{cases} 
1 & \text{if } \mathbf{\hat{y}} = \mathbf{y}, \\
0 & \text{otherwise}.
\end{cases}
\]

\textbf{Position-Wise (Hamming) Accuracy}
The proportion of correctly placed clips:
\[
\text{Hamming Accuracy} = \frac{1}{n} \sum_{i=1}^n \mathbb{I}(\hat{y}_i = y_i)
\]
where \( \mathbb{I}(\cdot) \) is the indicator function.

Additional metrics, such as Longest Common Subsequence (LCS) and Edit Distance (Levenshtein Distance), are presented in the Appendix \ref{sec:additional Metrics}. Due to the high cost of inference on this benchmark, which includes 22,846 clips for two modalities, we evaluated each model only once, as preliminary tests indicated minimal variation across multiple runs.

%\[
%\text{Hamming Accuracy} = \frac{1}{n} \sum_{i=1}^n \mathbb{I}(\hat{y}_i = y_i)
%\]
%where \( \mathbb{I}(\cdot) \) is the indicator function.

%\[
%\text{Binary Accuracy} = 
%\begin{cases} 
%1 & \text{if } \mathbf{\hat{y}} = \mathbf{y}, \\
%0 & \text{otherwise}.
%\end{cases}
%\]

\subsection{Models}

Our benchmark requires models to process multiple videos and reference them accurately. Simply merging frames from different clips is not sufficient; the model must correctly identify and arrange them in sequence, a capability still uncommon in current state-of-the-art models. To assess this, we ran a sanity check by inputting multiple clips into candidate models to evaluate whether they could reference and describe them coherently. Based on these results, we selected Qwen2-VL-Instruct \citep{wang2024qwen2}, Gemini-Flash \citep{gemini20flash}, InternVL2.5 \citep{chen2024expanding}, and LlavaOnevision \citep{li2024llava} as they were among the few able to perform this task reliably. 

We also conducted preliminary ablation studies to determine the best frame-sampling strategies for different models. For Qwen2.5-VL, tests at 1, 2, and 3 fps showed minimal differences. InternVL2.5 was tested at 4, 8, 16, and 32 frames; we found that 16 frames yielded better performance than lower counts, and no significant improvement was noted beyond 16 frames. Thus, we selected 16 frames for models like InternVL2.5 and LlavaOneVision, which have uniform frame distribution as default settings, and 1 fps for models defaulting to fps. Full test settings, prompts, and model details are provided in Appendices \ref{sec:test-settings} and \ref{sec:prompts}.

\begin{figure}[t]
    \begin{center}
        \centerline{
            \begin{minipage}{\linewidth}
                \centering
                \includegraphics[width=\linewidth]{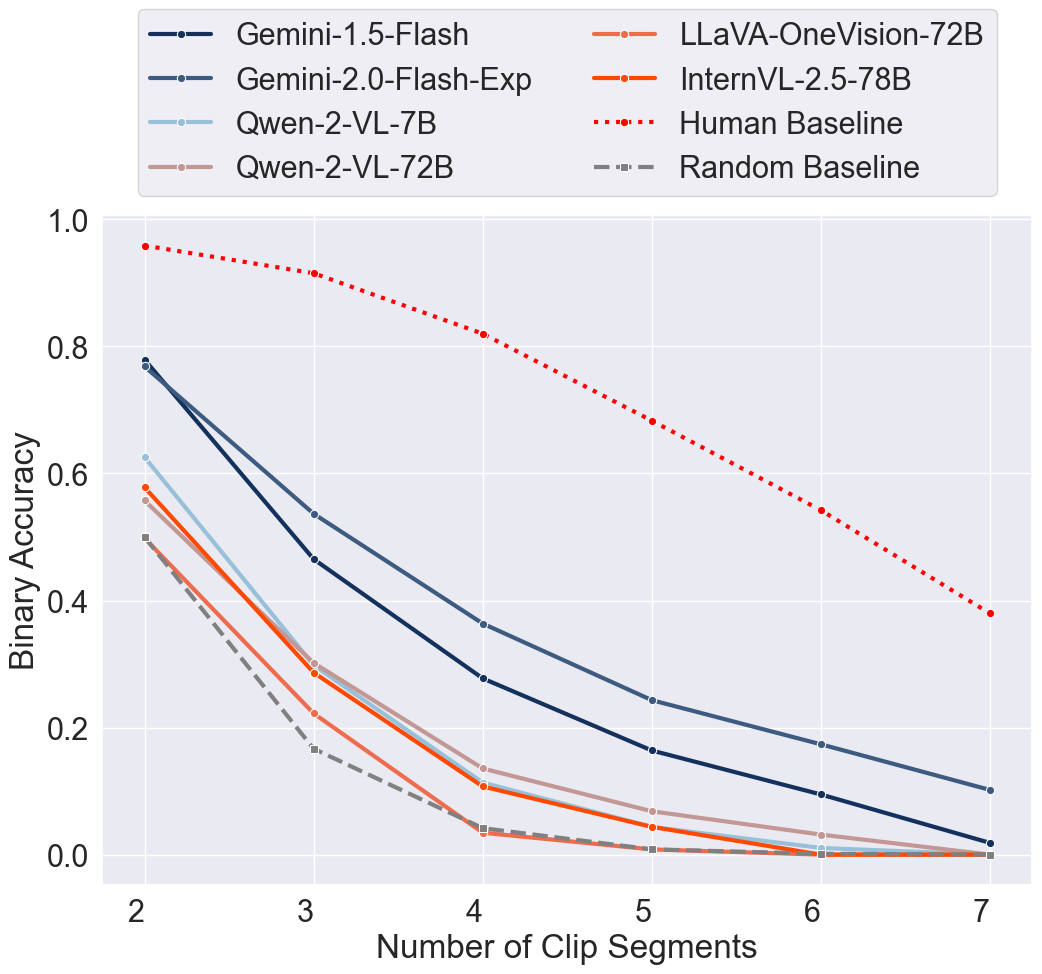}
            \end{minipage}
        }
        \caption{Binary accuracy versus the number of clips (2–7), comparing various state-of-the-art VLMs against human and random baselines}
        \label{fig:square_single_plot}
    \end{center}
    \vskip -0.2in
\end{figure}

\begin{table*}[t]
\centering
\begin{tabular}{l | cc | cc | cc}
\hline
  & \multicolumn{2}{c|}{Vision Only} & \multicolumn{2}{c|}{Vision+Text} & \multicolumn{2}{c}{Text} \\
  & Binary & Hamming & Binary & Hamming & Binary & Hamming \\
\hline
Random           & 0.2114 & 0.3385  & 0.2114 & 0.3385 & -- & -- \\
\hline\hline
Human            & 0.8486 & 0.8855  & 0.8332 & 0.8904 & -- & -- \\
\hline\hline
Qwen2-VL-7B         & 0.3091 & 0.4432  & 0.4354 & 0.5683 & 0.3318  & 0.4924 \\
Qwen2-VL-72B        & 0.2990 & 0.4170  & 0.5708 & 0.6820 & \textbf{0.5402} & \textbf{0.6907} \\
Gemini-1.5-Flash    & 0.4599 &  0.5825 &   0.5936 & 0.7115 & 0.4642 & 0.6029 \\
Gemini-2.0-Flash-Exp & \textbf{0.5108} & \textbf{0.6188}  & \textbf{0.6939} & \textbf{0.7931} & 0.5271 & 0.6652 \\
InternVL2.5-78B     & 0.2899 &  0.4243 & 0.4856 & 0.6046    & 0.4768 & 0.6062 \\
LLaVA-OneVision-72B  & 0.2260 &  0.3514 &  0.4256      &   0.5597     & 0.4210 & 0.5545 \\
\hline
\end{tabular}
\caption{Binary and Hamming accuracy scores for various VLMs across different input modalities: Vision Only, Text Only, and Vision+Text. Human and random baselines are included for comparison.}
\label{tab:model_comparison}
\end{table*}

\section{Results}

The performance of the models compared to humans is shown in Table \ref{tab:model_comparison}, with different input settings and metrics, along with random accuracy calculated based on the number of clips in the videos. In terms of binary accuracy, where a prediction is considered correct only if it exactly matches the ground truth, humans score 84.86\%, while the highest-performing model, Gemini-2.0-Flash-Exp, scores 51.08\%, and random accuracy is 21.14\% when using video-only input. This demonstrates that although the model performs well above random accuracy, it still lags behind human performance. In contrast, while human performance did not benefit from the additional text modality, models showed substantial improvement.

Even on videos that humans misordered, models rarely outdo them. For instance, Qwen2-VL-72B solved none of the 57 seven-clip videos humans got wrong, while Gemini-2.0-Flash-Exp solved only three. Likewise, out of 77 six-clip videos misordered by humans, Qwen2-VL fixed two, and Gemini-2.0-Flash-Exp five, highlighting the persistent gap in visual reasoning.

\section{Discussion}

\subsection{Models Performance}

The results indicate that open-source models still lag behind closed-source models like Gemini, particularly in visual reasoning tasks. However, this performance gap narrows when text input is introduced or in text-only evaluations, where Qwen2-VL-72B outperforms Gemini. Notably, Qwen2-VL-7B performs on par with Qwen2-VL-72B in visual reasoning, suggesting that increasing the language model size does not specifically enhance visual reasoning capabilities, given that both models utilize the same vision encoder. Furthermore, SPLICE proves to be a particularly challenging benchmark, as models like LlavaOneVision perform at random levels, while InternVL-2.5-78B scores just above random chance with video only settings. 

\subsection{Human Performance}

Human performance on the binary metric reaches around 84\%, reflecting the challenge of tasks requiring commonsense and domain-specific knowledge. This ceiling highlights the difficulty of instructional videos, where distinguishing sub-steps in technical domains like automotive or medical tasks depends on expert knowledge. Notably, performance did not improve with text, likely because the short captions (averaging 4.84 words) rarely add information beyond the video. As a result, performance remains consistent across metrics.

\subsection{Input Modality}

In our study, we report results across three modalities: video-only, text-only, and video+text. However, the text-only results are not intended to assess the reasoning capabilities of language models, as the dataset was curated with a focus on video and video-text inputs. The text often lacks sufficient detail on its own and is primarily included to evaluate the impact of combining modalities. When paired with video, the descriptions become clearer through visual context. Thus, we use text-only performance solely as a baseline to measure gains from incorporating visual information.

\begin{figure*}[ht]
    \vskip 0.2in
    \begin{center}
        \includegraphics[width=\linewidth]{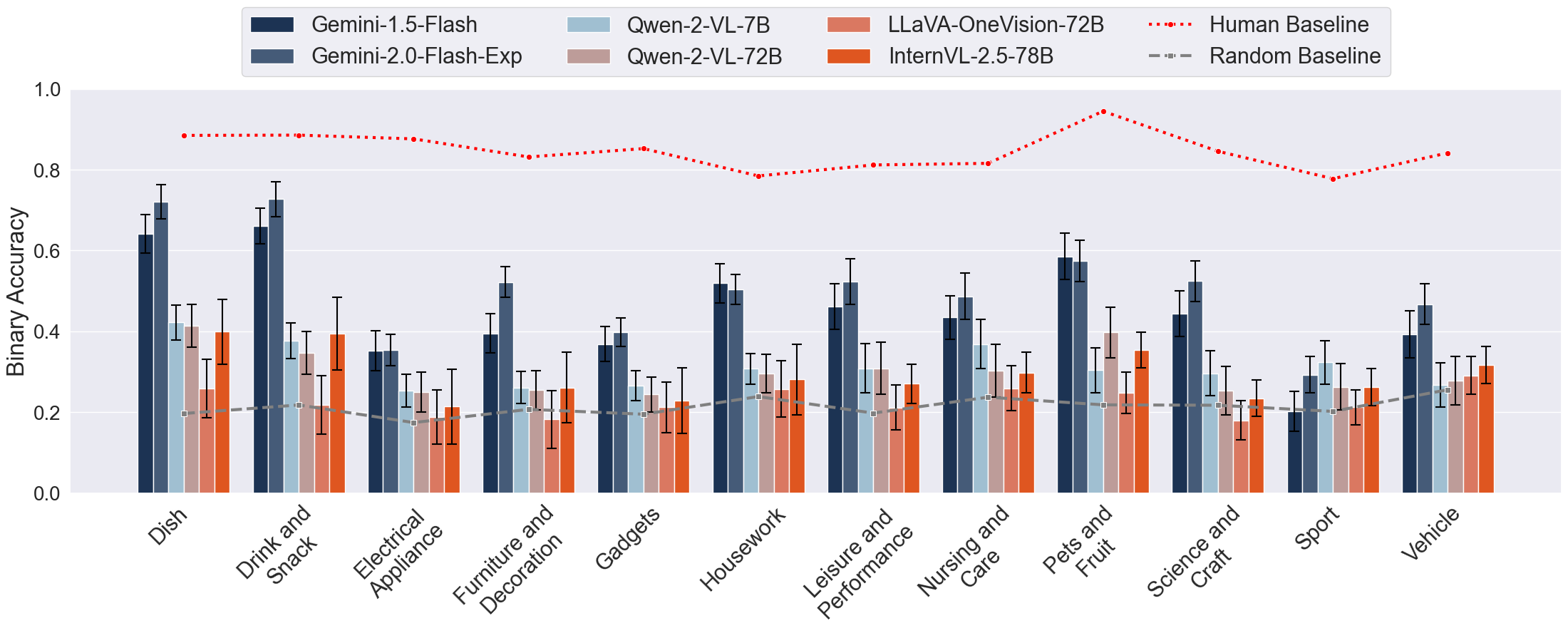}
        \vskip -0.06in
        \includegraphics[width=\linewidth]{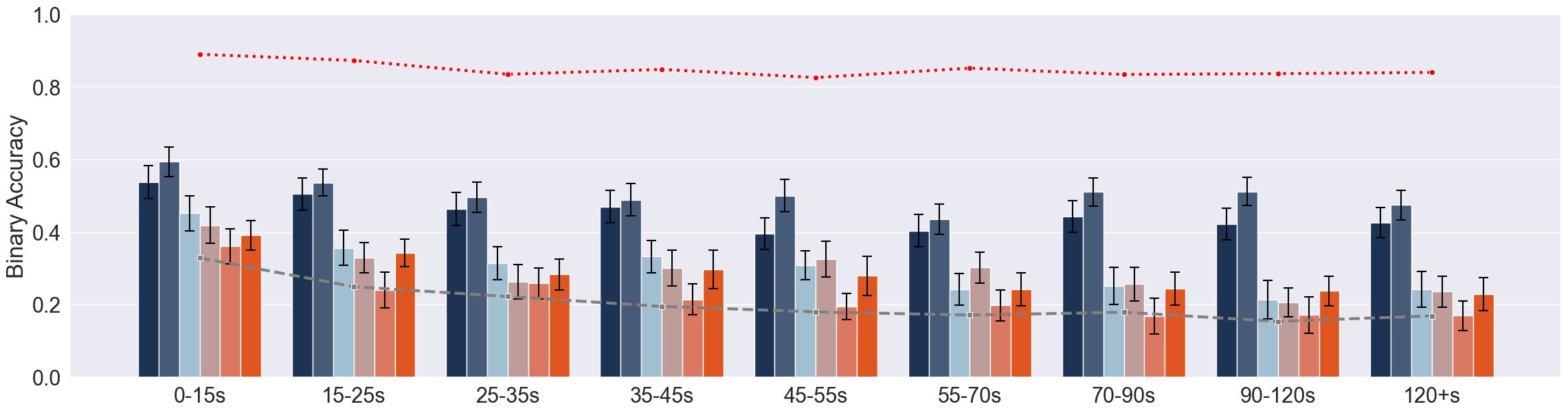}
        \caption{Binary accuracy performance of various state-of-the-art VLMs across different domains (top) and video durations (bottom), compared to a human baseline (red dashed line) and a weighted random baseline (gray dashed line). Error bars represent the 95\% confidence interval (CI).}
        \label{fig:main_results_double}
    \end{center}
    \vskip -0.2in
\end{figure*}

The results show that models benefit significantly from the additional text modality, unlike humans. This could be due to two factors. First, the text is a human-generated summary of what is happening in the video, meaning there is a form of knowledge distillation from humans to the model. In contrast, with video-only input, the model receives no additional human-provided context. Additionally, this improvement suggests that language models are still more capable of reasoning compared to Vision-Language Models (VLMs), as they were able to benefit from information that was not necessary for humans to correctly order the videos.

\subsection{Factors Influencing Performance}

One clear factor influencing the performance of both models and humans is the number of clips. As shown in Figure \ref{fig:square_single_plot}, performance declines for both groups as the number of clips to be ordered increases. However, models are more adversely affected than humans by this increase, and the benefit of including human annotations diminishes with longer sequences as shown in Figure \ref{fig:Appendix3} in the appendix. This trend aligns with findings from prior work \citep{sharma-etal-2024-losing}. As a result, most models perform near random when ordering seven clips, where the random-chance baseline is much lower due to the 7! possible permutations.

Figure \ref{fig:main_results_double} (left) shows varying performance across categories that does not always concur with human outcomes. For instance, electrical appliances is hard for most models but easy for humans, while both struggle with sports. Models in the same family typically show similar trends, but cross-family comparisons can differ. 
Another interesting finding illustrated is that Gemini, similar to humans, is not affected by the duration of the videos for both video only input (Figure \ref{fig:main_results_double} right) and video+text (Appendix, Figure \ref{fig:Appendix2}). It maintains stable performance even on longer videos, whereas Qwen2-VL-72B, and to an even greater extent Qwen2-VL-7B, exhibit performance degradation as video length increases. This suggests that Gemini has robust performance across long contexts.

\subsection{In-depth Analysis of Reasoning in Models}
\label{depthreasoning}
To better understand model failures, we analyzed sub-domains within the 12 main COIN domains. For example, the "sport" domain includes subclasses like "Practice Skiing Aerials" and "Practice Karate." Our analysis used only the video modality.

\begin{table}[th] % <<< Table environment starts INSIDE document
  \centering % <<< Caption INSIDE table
   % <<< Label INSIDE table (usually after caption)
  \begin{tabular}{l c}
    \hline
    \textbf{Models} & \textbf{Accuracy (\%)} \\
    \hline
    % --- Group: change/replace ---
    \multicolumn{2}{c}{\textbf{Group: change/replace, 328 videos}} \\
    %    \multicolumn{2}{c}{\small Videos: 328, Avg GT length: 3.74} \\
    \hline
    Random        & 15.02 \\
    Human         & 88.11 \\
    Gemini-1.5-Flash & 26.22  \\
    Gemini-2.0-Flash-Exp & 26.83  \\
    Qwen2-VL-72B  & 21.04 \\
    \hline
    % --- Group: other ---
    \multicolumn{2}{c}{\textbf{Group: other, 157 videos}} \\
    %    \multicolumn{2}{c}{\small Videos: 157, Avg GT length: 3.18} \\
    \hline
    Random        & 22.68 \\
    Human    & 86.62 \\
    Gemini-1.5-Flash & 54.14   \\
    Gemini-2.0-Flash-Exp & 53.50  \\
    Qwen2-VL-72B & 33.12  \\
    \hline
  \end{tabular}
\caption{Video Ordering Accuracy of Electrical Appliance domain for sub-domains that include change/replace compared to others that don't}
\label{tab:video_ordering_electrical}
\end{table} % <<< Table environment ends INSIDE document

\textbf{Contextual vs. Temporal Reasoning}
We first examined the "electrical appliance" domain, which exhibited the second-lowest model performance and contained 485 of the dataset's 3381 videos, to understand the reasons for this low performance. A closer look revealed that approximately 70\% of the videos involve actions like "replacing" or "changing". In these "change/replace" videos, the initial and final states are often visually similar. For example, in the "replace toner cartridge" sub-domain, a typical sequence involves opening the printer door, removing the old cartridge, inserting the new one, and closing the door. Consequently, the first and last clips (door closed) are visually similar to each other, as are the intermediate clips involving the cartridge exchange. Therefore, within the "electrical appliance" domain, we compared sub-domains involving "change/replace" actions against those without. We found a significant performance disparity: Gemini-2.0-Flash scored 26.83\% on "change/replace" videos but 53.5\% on other videos within the same domain. Results for all models are shown in Table \ref{tab:video_ordering_electrical}. While models performance drops sharply for the "change/replace" group, human performance remains largely unaffected.

To investigate this performance gap further, we examined the models' predictions for "change/replace" videos. We found that models frequently predicted the first and last clips as being sequential. For instance, in videos with four or more clips, Gemini-2.0 predicted the first and last clips were adjacent 57.36\% of the time. In contrast, human annotators made this prediction only 2.45\% of the time, while random chance would be 27.29\%. This indicates a model bias towards ordering visually similar clips adjacently. We attribute this to a potential lack of contextual understanding, where models take shortcuts by associating visually similar clips as sequential rather than performing deeper contextual reasoning (e.g., recognizing that opening a door precedes intermediate actions and closing the door comes last). However, in these same videos, models like Gemini-2.0-Flash demonstrated relatively better temporal reasoning concerning the first and last clips (despite their visual similarity), correctly identifying the first clip as preceding the last about 70\% of the time.

\textbf{Contextual vs. Causal Reasoning}
We further investigated this hypothesis across all domains using a larger set of videos, filtered into two groups: "make" and "change/replace". Our hypothesis was that "make" videos primarily require causal and temporal reasoning, whereas "change/replace" videos, as previously argued, depend more heavily on contextual reasoning. For example, in "make" tasks like preparing food, mixing ingredients causally changes the food's state; similarly, when making a bed, folding the covers has a causal relationship to the finished state. The dataset contains 844 "change/replace" videos (average 3.57 clips) and 765 "make" videos (e.g., make ice cream, make bed; average 3.53 clips), indicating similar complexity based on clip count and thus a comparable random baseline. The results in Table \ref{tab:video_ordering_combined} reveal a significant performance difference between the "change/replace" and "make" groups, suggesting models possess stronger causal reasoning abilities compared to contextual reasoning with models like Gemini-2.0-Flash-Exp scoring 32.11\% on "change/replace" videos, but 68.37\% on "make videos".

\begin{table*}[t]
\centering
% NOTE: Caption IS inside the table* environment

\begin{tabular}{l | c c || c c || c}
\toprule % Use booktabs rule for better spacing
\textbf{Models} & \multicolumn{2}{c||}{\textbf{Causal vs. Contextual}} & \multicolumn{2}{c||}{\textbf{General Knowledge}} & \textbf{Spatial} \\
\cmidrule(lr){2-3} \cmidrule(lr){4-5} % Partial lines for subheadings
 & \textbf{Change/Replace} & \textbf{Make} & \textbf{Everyday} & \textbf{Technical} & \\
\midrule % Use booktabs rule
Number of videos     & 844  & 765    & 342    & 300  & 138  \\
\midrule
Random        & 18.02  & 19.14    & 23.26  & 22.49  & 19.19  \\
\midrule % Separator line
Human         & 84.95  & 86.01 & 85.38 & 80.00 & 76.09   \\
Gemini-1.5-Flash & 28.44  & 61.18 & 56.14 & 44.33 & 24.64  \\
Gemini-2.0-Flash-Exp & 32.11  & 68.37 & 59.65 & 47.00 & 36.96 \\
Qwen2-VL-72B  & 20.38  & 36.86 & 36.55 & 32.00 & 21.01  \\
\bottomrule % Use booktabs rule
\end{tabular}
\caption{Models and human video ordering accuracy (\%) on task subsets probing different reasoning types}
\label{tab:video_ordering_combined}
\end{table*}

\textbf{General Knowledge Reasoning}

Next, we defined two groups: "Everyday tasks" and "Specialized/Technical tasks". "Everyday tasks" include activities like washing dishes or ironing clothes, while "Specialized/Technical tasks" involve actions such as drawing blood or installing a ceiling fan. To avoid earlier issues with "change/replace" scenarios, we excluded such videos from both groups. This resulted in 342 videos labeled as "Everyday tasks" and 320 as "Specialized/Technical tasks". Results in Table \ref{tab:video_ordering_combined} show that both models and humans perform better on "Everyday tasks". These tasks likely rely more on commonsense reasoning and are better represented in the models' training data. The performance drop on specialized tasks suggests limitations in the models' knowledge and reasoning in less familiar technical contexts.

\textbf{Spatial Reasoning}
Finally, we identified a set of sub-domains requiring spatial reasoning. Examples include "Practice Skiing Aerials", where the model must recognize that the skier in the air precedes landing; pole vaulting, involving jumping, being airborne, and landing; and weightlifting, where the weight on the ground precedes being lifted. Results indicate that models struggle with these spatial reasoning sub-domains; for instance, Gemini-2.0-Flash achieved only 36.96\% accuracy, compared to its overall dataset average of 51.08\% as shown in Table \ref{tab:video_ordering_combined}. This likely explains the lower performance observed in the "sport" domain, as 8 out of its 10 sub-domains heavily involve spatial reasoning.

\section{Conclusion}

In this paper, we presented a novel, human-curated benchmark designed to assess multiple facets of visual reasoning, including temporal, causal, contextual, visual-spatial, and general knowledge. We evaluated various open-source and closed-source VLMs under different input modalities and compared their performance against human participants. Despite improvements from combining video and text (highlighting the value of cross-modal alignment), all models lag significantly behind human performance, especially without human-annotated descriptions. Moreover, open-source models lag further behind their closed-source counterpart, revealing a persistent gap in visual reasoning. The low performance of several models, with some scoring just above random chance, highlights the benchmark’s effectiveness as a rigorous probe of visual reasoning. Our reasoning analysis further reveals that models perform better on videos dominated by causal and temporal reasoning than on those requiring contextual or spatial reasoning. In the future, we aim to incorporate voice to enhance cross-modal alignment and assess how models integrate audio with visual reasoning.

\section{Limitations}

Currently, only a few state-of-the-art VLMs support the ability to input multiple videos and reference them appropriately. Even single-video processing capabilities are limited in many models, restricting our evaluation to the handful that do offer this functionality. Nonetheless, the field is evolving rapidly, and we expect that most models will soon be able to handle multi-video inputs, enabling broader application of our benchmark.

\section{Acknowledgements}
This work was funded by the Deutsche Forschungsgemeinschaft (DFG, German Research Foundation) - 456666331


%\bibliography{custom}

\newpage

\appendix

\onecolumn
\section{Appendix}

\subsection{Additional Metrics and full results}
\label{sec:additional Metrics}

Each video in the dataset is segmented into clips based on the original COIN dataset's step localization. These clips are then randomly shuffled and renamed as \( C = \{c_1, c_2, \dots, c_n\} \). Depending on the modality being tested, the model receives either video-only input or a combination of video and annotations (short textual descriptions of events in the video). The model's task is to predict the correct permutation of the clip order. Let \( \mathbf{y} = [y_1, y_2, \dots, y_n] \) denote the ground-truth sequence of clip indices, and \( \mathbf{\hat{y}} = [\hat{y}_1, \hat{y}_2, \dots, \hat{y}_n] \) represent the predicted sequence. For \( n \) clips, there exist \( n! \) possible permutations. The models are evaluated on four metrics:

\textbf{Binary Accuracy}. The prediction is correct only if the entire sequence matches the ground truth:
\[
\text{Binary Accuracy} = 
\begin{cases} 
1 & \text{if } \mathbf{\hat{y}} = \mathbf{y}, \\
0 & \text{otherwise}.
\end{cases}
\]

\textbf{Position-Wise (Hamming) Accuracy}
The proportion of correctly placed clips:
\[
\text{Hamming Accuracy} = \frac{1}{n} \sum_{i=1}^n \mathbb{I}(\hat{y}_i = y_i)
\]
where \( \mathbb{I}(\cdot) \) is the indicator function.

\textbf{Longest Common Subsequence (LCS)}.
The LCS measures the longest sequence of elements appearing in the same relative order in both \( \mathbf{y} \) and \( \mathbf{\hat{y}} \). Let \( c(i, j) \) denote the length of the LCS for substrings \( \mathbf{y}_{1:i} \) and \( \mathbf{\hat{y}}_{1:j} \):
\[
\small
c(i, j) = 
\begin{cases} 
0 & i = 0 \text{ or } j = 0, \\
c(i-1, j-1) + 1 & y_i = \hat{y}_j, \\
\max\{c(i-1, j), c(i, j-1)\} & \text{otherwise}.
\end{cases}
\]
The LCS ratio normalizes this value:
\[
\text{LCS Ratio} = \frac{\text{LCS Length}}{n}
\]

\textbf{Edit Distance (Levenshtein Distance)}.
The minimum number of insertions, deletions, or substitutions required to transform \( \mathbf{\hat{y}} \) into \( \mathbf{y} \). Define a matrix \( D \) where \( D(i, j) \) is the edit distance between \( \mathbf{y}_{1:i} \) and \( \mathbf{\hat{y}}_{1:j} \):
\[
D(i,0) = i, \quad D(0,j) = j \quad \text{(boundary conditions)},
\]
\[
\small
D(i,j) = 
\begin{cases}
D(i-1,j-1) & y_i = \hat{y}_j, \\
1 + \min\left(
\begin{aligned}
&D(i-1,j) \\ 
&D(i,j-1) \\ 
&D(i-1,j-1)
\end{aligned}
\right) & \text{otherwise}.
\end{cases}
\]
The final edit distance is \( D(n, n) \).

\begin{table*}
\centering
\small
\begin{tabular}{c | l | cccc|cccc}

\hline
& & \multicolumn{4}{c|}{Vision Only} & \multicolumn{4}{c}{Vision+Text} \\
\cline{3-10}
& & Binary & Hamming & LCS & Edit & Binary & Hamming & LCS & Edit \\
\hline
& Random & 0.2114 & 0.3385 & 0.6554 & 2.0970  & 0.2114 & 0.3385 & 0.6554 & 2.0970  \\
\hline\hline
& Human & 0.8486 & 0.8855 & 0.9359 & 0.4105 & 0.8332 & 0.8904 & 0.9337 & 0.3875 \\
\hline\hline
& Qwen2-VL-7B  & 0.3091  & 0.4432 & 0.7130 & 1.7891 & 0.4354 & 0.5683 & 0.7896 & 1.4377 \\
& Qwen2-VL-72B  & 0.2990 & 0.4170 & 0.7011 & 1.8465 & 0.5708 & 0.6820 & 0.8483 & 1.0677 \\
& Gemini-1.5-Flash & 0.4599 &  0.5825& \textbf{0.7980} &  1.3458 &   0.5936 & 0.7115 & 0.8633 & 0.9734 \\
& Gemini-2.0-Flash-Exp & \textbf{0.5108 } & \textbf{0.6188} & 0.7927 & \textbf{1.2511} &  \textbf{0.6939} & \textbf{0.7931} & \textbf{0.9030} & \textbf{0.7788} \\
& InternVL2.5-78B   &  0.2899 & 0.4243 & 0.7050 & 1.8364 & 0.4856 & 0.6046 & 0.7694 & 1.3602 \\
& LlavaOnevision-72B  & 0.2260 & 0.3514 & 0.6615 &  2.0636 & 0.4256 & 0.5597 & 0.7866 & 1.4312 \\
\hline
\end{tabular}
\caption{Performance comparison of various VLMs across different input modalities (Vision Only and Vision+Text) using Binary Accuracy, Hamming Distance, Longest Common Subsequence (LCS), and Edit Distance metrics. Human and random baselines are included for reference. Models perform significantly better with textual input, highlighting the benefit of cross-modal information}
\label{tab:model_comparison2}
\end{table*}

\newpage

\subsection{Testing Settings}
Below are the details about test settings of each model:

\label{sec:test-settings}
\textbf{Qwen2-VL-Instruct Family}. Qwen2-VL was tested with both 7B and 72B parameters. The number of frames was set to 1 fps, and the highest image resolution was set to 448 pixels, while the other dimension was automatically adjusted based on the aspect ratio of the input frames.

\textbf{Gemini-Flash Family}. We used Gemini Flash 1.5 and 2.0 (experimental) versions, with the fps set to 1. The model was loaded using the official Google API, and the image resolution was left at the default setting, allowing the model to handle it automatically. 

\textbf{InternVL2.5 Family}. InternVL2.5 was tested with the 78B parameters model only. The 8B model did not pass the sanity check. We used the default settings of the uniformal distribution of frames input for each clip and we set it to 16 frames instead of fps. 

\textbf{LlavaOnevision Family} LlavaOnevision was tested with 72B parameters. The 7B model did not pass the sanity check. We used the default settings of the uniformal distribution of frames input for each clip and we set it to 16 frames instead of fps. 

All of the open source models were used from the Hugging Face library \citep{wolf_huggingfaces_2019} and adopted with the Flash Attention approach.
All of these models are tested with three different modalities, vision only, text only, and vision + text. Samples of the prompts are shown in the Appendix \ref{sec:prompts}. All jobs were submitted to a cluster of A100 and H100 GPUs, which were used interchangeably based on availability.

\newpage
\subsection{Prompts}
Three samples of prompts are shown below, for each model the prompts were slightly tuned for better performance:

Here is a sample prompt for video-only input:
\textbf{prompt:}  f"A video has been split into {len(clips)} clips, shuffled randomly."
"Your task is to analyze each clip deeply to reorder them into the correct temporal sequence. Focus on:"
"1. Visual content: Examine the actions, transitions, scene details, and context within each clip."
"Provide the reordered sequence strictly within order tags in this format: "
"'<order>Video X, Video Y, Video Z, ...</order>'." 

Here is a sample prompt for video+text input:
\textbf{prompt:}  f"A video has been split into {len(clips)} clips, shuffled randomly."
                "Your task is to analyze each clip deeply to reorder them into the correct temporal sequence. Focus on:"
                "1. *Visual content*: Examine the actions, transitions, scene details, and context within each clip."
                "2. *Temporal logic*: Identify the logical progression of events based on what happens before or after."
                "3. *Annotations*: Leverage the annotations to infer their proper chronological sequence."
                "Provide the reordered sequence strictly within order tags in this format: "
                "'<order>Video X, Video Y, Video Z, ...</order>'." 
                
Here is a sample prompt for text models:
\textbf{prompt:} 
        "Analyze the following video clips descriptions and order them chronologically as they are part of one continuous video. "
        "Focus on temporal clues, event progression, scene transitions and other cues "
        "Each video clip is labeled as ‘Video X’, where Video X corresponds to one shuffled clip. "
        "Maintain these labels in your response. "
        
        "Return the ordered video strictly within <order> tags in this format: "
        "<order>Video X, Video Y ...</order> "
    )
    
\label{sec:prompts}

\newpage

\subsection{Videos Stats}
\label{sec:vid-stats}

Table \ref{tab:segment_distribution} provides statistics on the segmented video dataset, detailing how videos are divided into segments and their distribution across different segment counts.

\begin{table}[ht]
  \centering
  \resizebox{1\columnwidth}{!}{
  \begin{tabular}{cccccccccccc}
    \hline
    \textbf{Segments} & \textbf{Videos} & \textbf{Clips} & \textbf{Mean Time (s)} & \textbf{Std Dev} & \textbf{(2, 35]} & \textbf{(35, 68]} & \textbf{(68, 100]} & \textbf{(100, 133]} & \textbf{(133, 166]} & \textbf{(166, 198]} & \textbf{(198, 330]} \\
    \hline
    2 & 1020 & 2040 & 46.84 & 40.75 & 531 & 245 & 128 & 65 & 35 & 12 & 4 \\
    3 & 1026 & 3078 & 53.32 & 42.68 & 434 & 309 & 142 & 79 & 36 & 19 & 7 \\
    4 & 734 & 2936 & 62.41 & 40.76 & 209 & 260 & 146 & 70 & 32 & 13 & 4 \\
    5 & 333 & 1665 & 72.67 & 43.20 & 62 & 123 & 69 & 46 & 21 & 10 & 2 \\
    6 & 172 & 1032 & 67.86 & 37.49 & 38 & 56 & 45 & 22 & 9 & 2 & 0 \\
    7 & 96 & 672 & 73.50 & 38.28 & 19 & 26 & 30 & 14 & 5 & 2 & 0 \\
    \hline
  \end{tabular}
  }
  \caption{This table summarizes the distribution of videos based on their segmentation. It includes the number of segments(2-7), total videos per segment number, total clips, mean  duration (seconds), and standard deviation. The rightmost columns show the distribution of videos across predefined video duration intervals, providing insights into the dataset’s temporal structure for event ordering analysis.}
  \label{tab:segment_distribution}
\end{table}

\newpage

\subsection{Instructions for Annotators}

\label{sec:instruct}

Instruction Brief 
Task: Reorder the video parts for each folder into their correct sequence.
Steps:
Download and Open the Folder assigned to you:
You will receive a folder containing several subfolders, each labeled with a unique number (e.g., 1, 2, 3, etc.).
Each subfolder corresponds to a video task with shuffled parts.
View the Video Parts:
Inside each subfolder, you will find video parts named random\_part\_1.mp4, random\_part\_2.mp4, etc.
These parts contain embedded labels as secondary context for your understanding of the video context.
Reorder the Parts:
Watch each video part carefully.
Determine the correct sequence of these parts based on the visual and textual cues.
Write down the sequence in the format:
Folder Number: Correct Order (e.g., 1: random\_part\_3, random\_part\_1, random\_part\_2). For simplicity use [2, 3, 4, 5, 1], where each number represents the Random number video.
Use “unk” in these cases:

1- Repeated instructions: If the video contains two separate instances of the same instruction.

2- Continuous actions without sufficient context: An action extends across multiple clips with insufficient background information to establish a clear sequence.

3- Unrelated actions: The video includes unrelated actions with no contextual clues to establish order.

Submit Your Results:
Compile the correct order for all folders in the attached spreadsheet
Use “unk” for any task sample you believe makes no sense or as discussed during the meeting, 
Notes:
Do not use any external sources 
Complete all tasks to the best of your ability.

\newpage
\subsection{Additional Results}

\newpage
\subsection{Additional Figures}
\subsubsection{Example of Video Input}
\begin{figure*}[th]
  \centering
  \includegraphics[width=\textwidth]{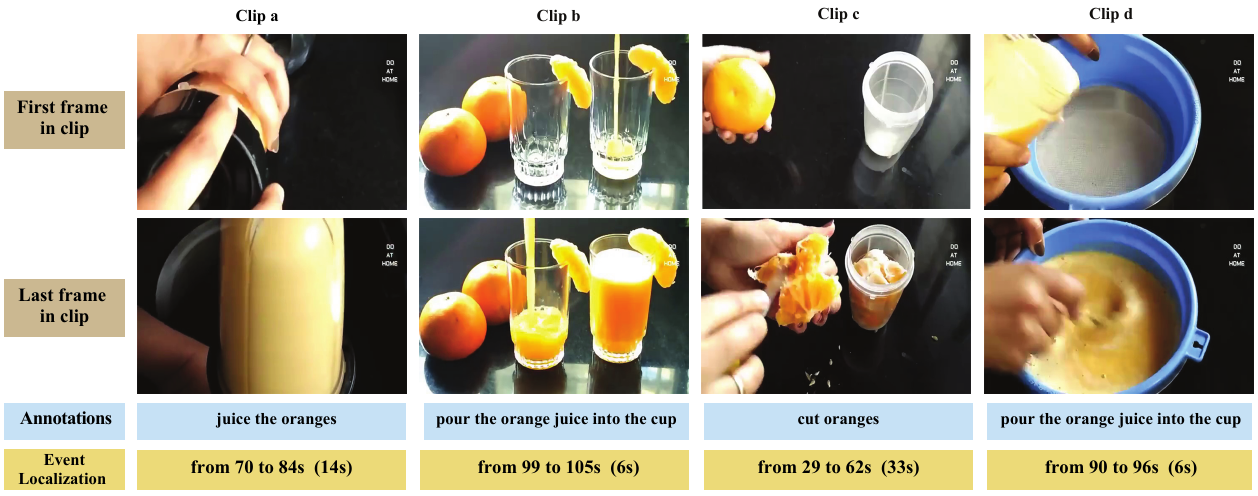} 
  \caption{An example of a set of clips that the models need to order correctly. The figures show the first and last frames of each clip. The clips are segmented based on events, reducing the reliance on shortcuts. Clip durations vary based on the event, with gaps where moments not relevant to the main steps are omitted. In this video, models must infer that the oranges are cut, juiced, filtered, and then served. Clips order: c, a, d, b. }
  \label{fig:sample}
\end{figure*}

\newpage

\subsubsection{Modalities Across Domains}
\begin{figure*}[ht]
    \vskip 0.2in
    \begin{center}
        \begin{subfigure}[b]{0.9\linewidth}
            \centering
            \includegraphics[width=\linewidth]{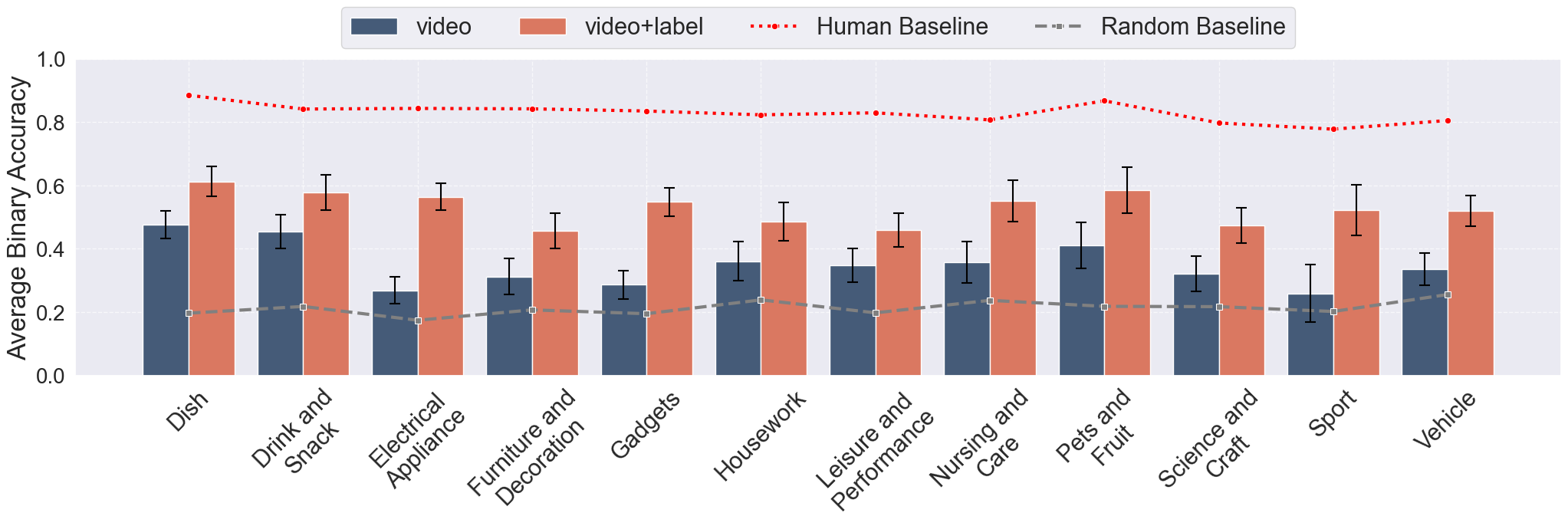}
            \caption{Average VLM performance on both modalities across video domains.}
            \label{fig:vlm_average_performance_by_domain}
        \end{subfigure}
        \vskip 0.1in
        
        \begin{subfigure}[b]{0.9\linewidth}
            \centering
            \includegraphics[width=\linewidth]{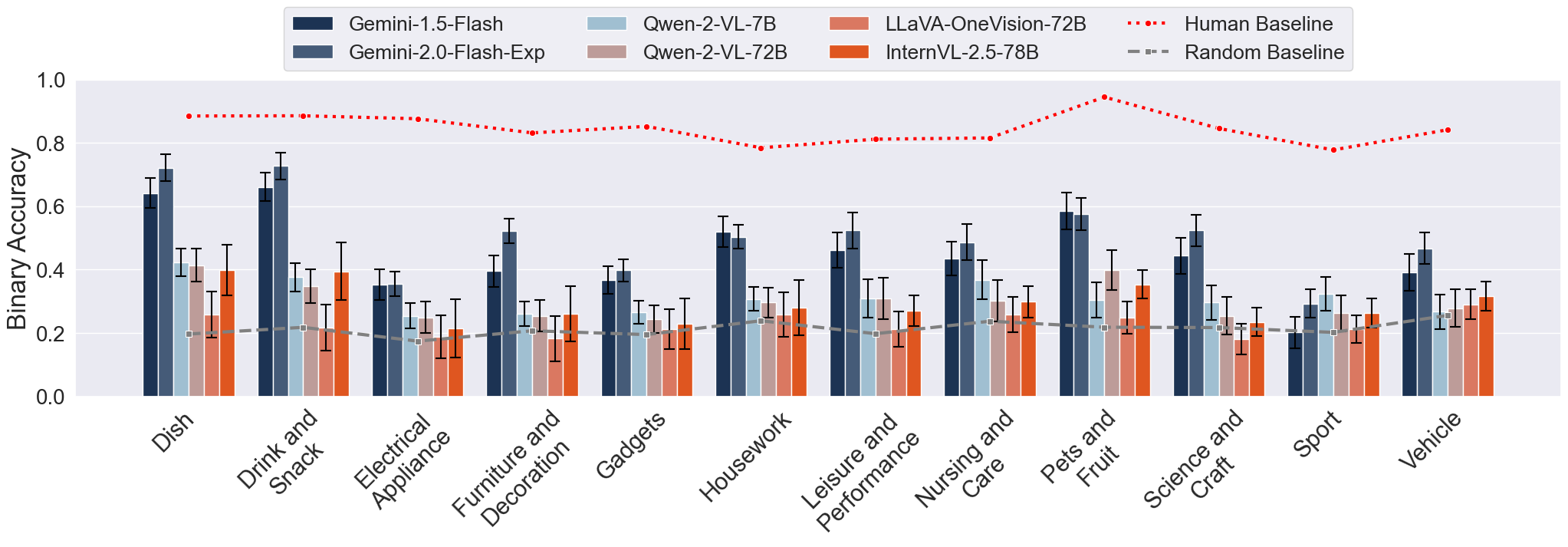}
            \caption{VLM performance by video domains with video only input}
            \label{fig:vlm_performance_by_domain}
        \end{subfigure}
        \vskip 0.1in
        
        \begin{subfigure}[b]{0.9\linewidth}
            \centering
            \includegraphics[width=\linewidth]{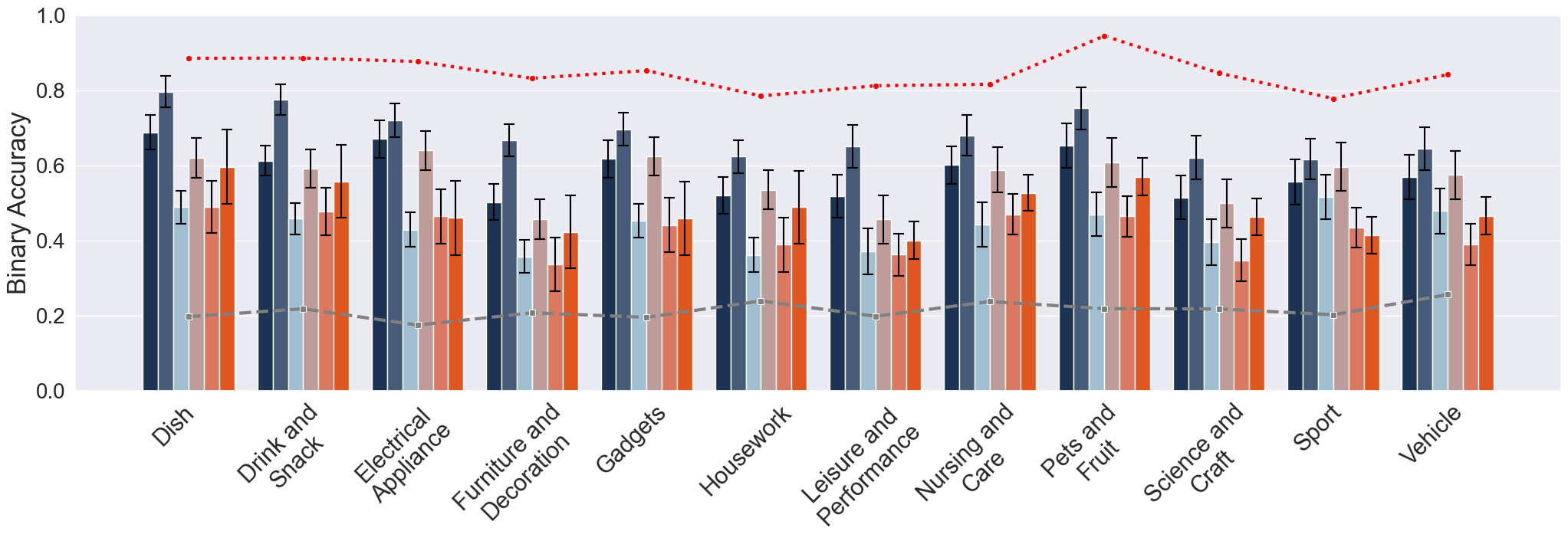}
            \caption{VLM performance by video domains with video and text input}
            \label{fig:vlm_performance_by_domain_labels}
        \end{subfigure}
        
        \caption{Binary accuracy performance of various state-of-the-art VLMs across different domains and modalities compared to a human baseline (red dashed line) and a weighted random baseline (gray dashed line). Error bars represent the 95\% confidence interval (CI).}
        \label{fig:Appendix1}
    \end{center}
    \vskip -0.2in
\end{figure*}

\newpage
\subsubsection{Modalities Across Video Length}
\begin{figure*}[ht]
    \vskip 0.2in
    \begin{center}
        \begin{subfigure}[b]{\linewidth}
            \centering
            \includegraphics[width=.9\linewidth]{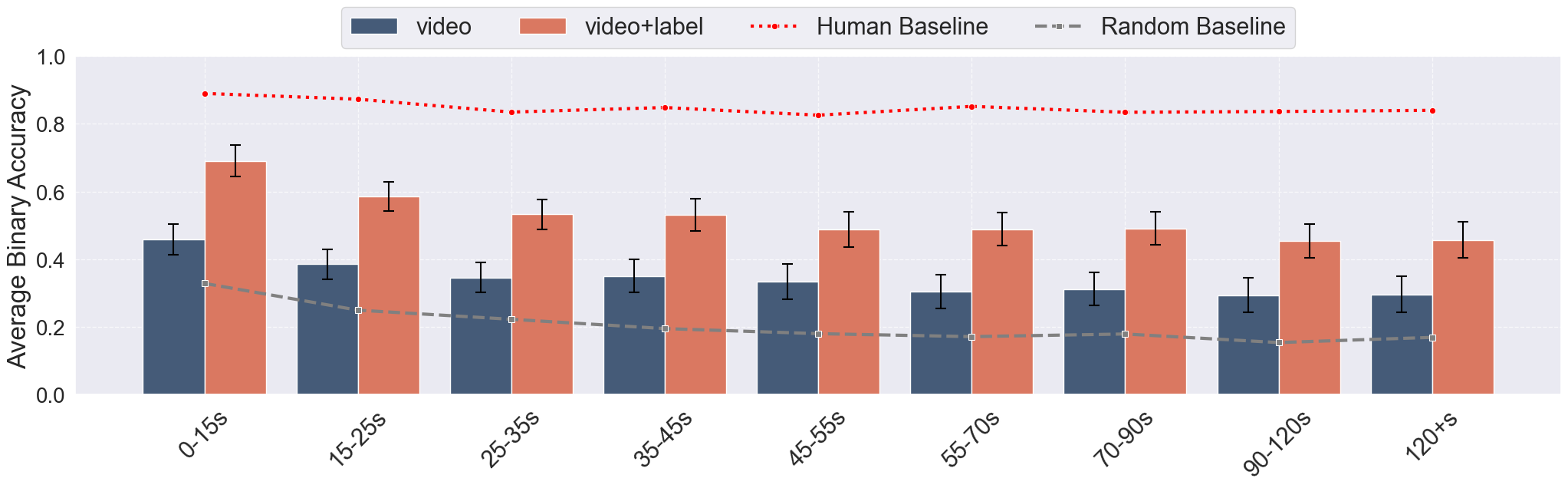}
            \caption{Average VLM performance on both modalities across video length.}
            \label{fig:vlm_performance_by_domain_labels}
        \end{subfigure}
        \vskip 0.1in
        
        \begin{subfigure}[b]{\linewidth}
            \centering
            \includegraphics[width=.9\linewidth]{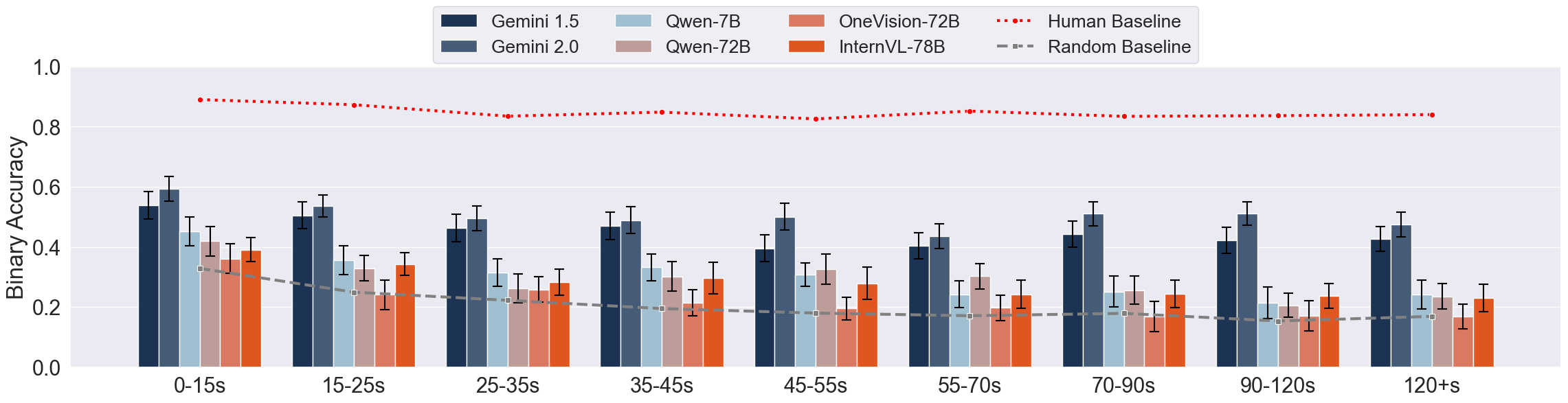}
            \caption{VLM performance by video duration with video only input}
            \label{fig:vlm_performance_by_duration_labels}
        \end{subfigure}
        \vskip 0.1in
        
        \begin{subfigure}[b]{\linewidth}
            \centering
            \includegraphics[width=.9\linewidth]{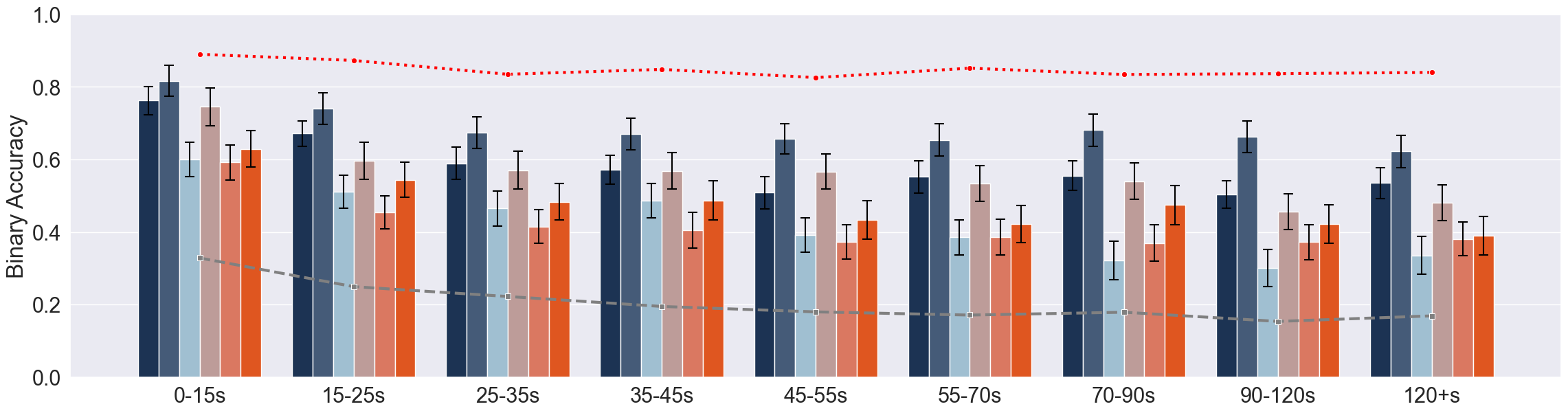}
            \caption{VLM performance by video duration with video and text input}
            \label{fig:vlm_performance_by_clip_length_labels}
        \end{subfigure}
        
        \caption{Binary accuracy performance of various state-of-the-art VLMs across video duration and modalities compared to a human baseline (red dashed line) and a weighted random baseline (gray dashed line). Error bars represent the 95\% confidence interval (CI).}
        \label{fig:Appendix2}
    \end{center}
    \vskip -0.2in
\end{figure*}

\newpage
\subsubsection{Modalities Across Number of Clips}
\begin{figure*}[ht]
    \vskip 0.2in
    \begin{center}
        \begin{subfigure}[b]{0.8\linewidth}
            \centering
            \includegraphics[width=\linewidth]{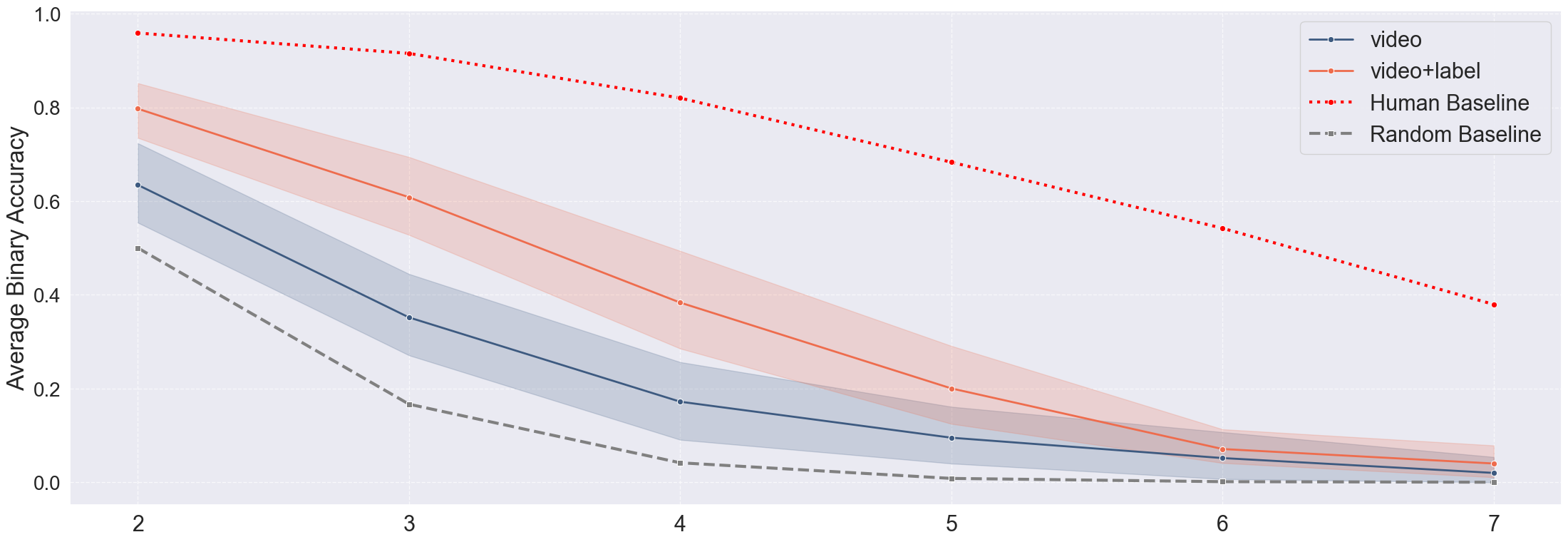}
            \caption{Average VLM performance on both modalities across number of Clips.}
            \label{fig:vlm_clip_length_performance}
        \end{subfigure}
        \vskip 0.1in
        
        \begin{subfigure}[b]{0.8\linewidth}
            \centering
            \includegraphics[width=\linewidth]{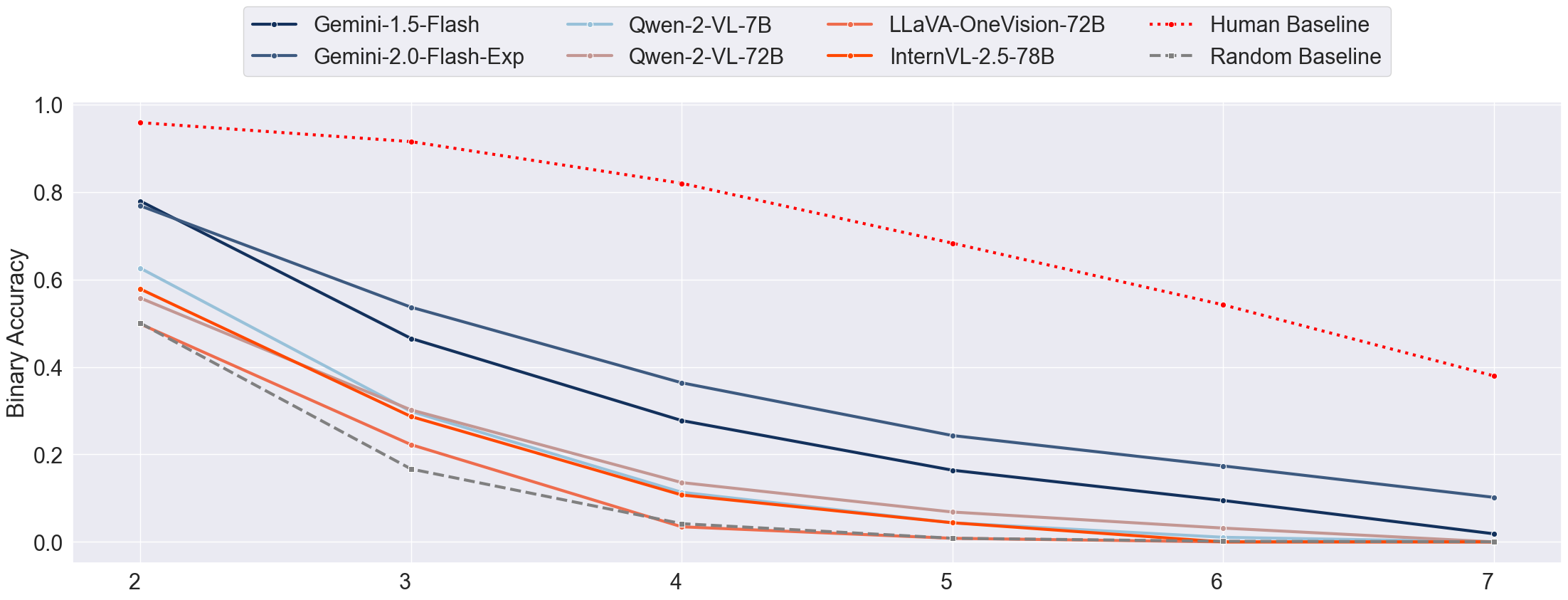}
            \caption{VLM performance comparison by number of clips with video only input}
            \label{fig:vlm_performance_by_clip_length}
        \end{subfigure}
        \vskip 0.1in
        
        \begin{subfigure}[b]{0.8\linewidth}
            \centering
            \includegraphics[width=\linewidth]{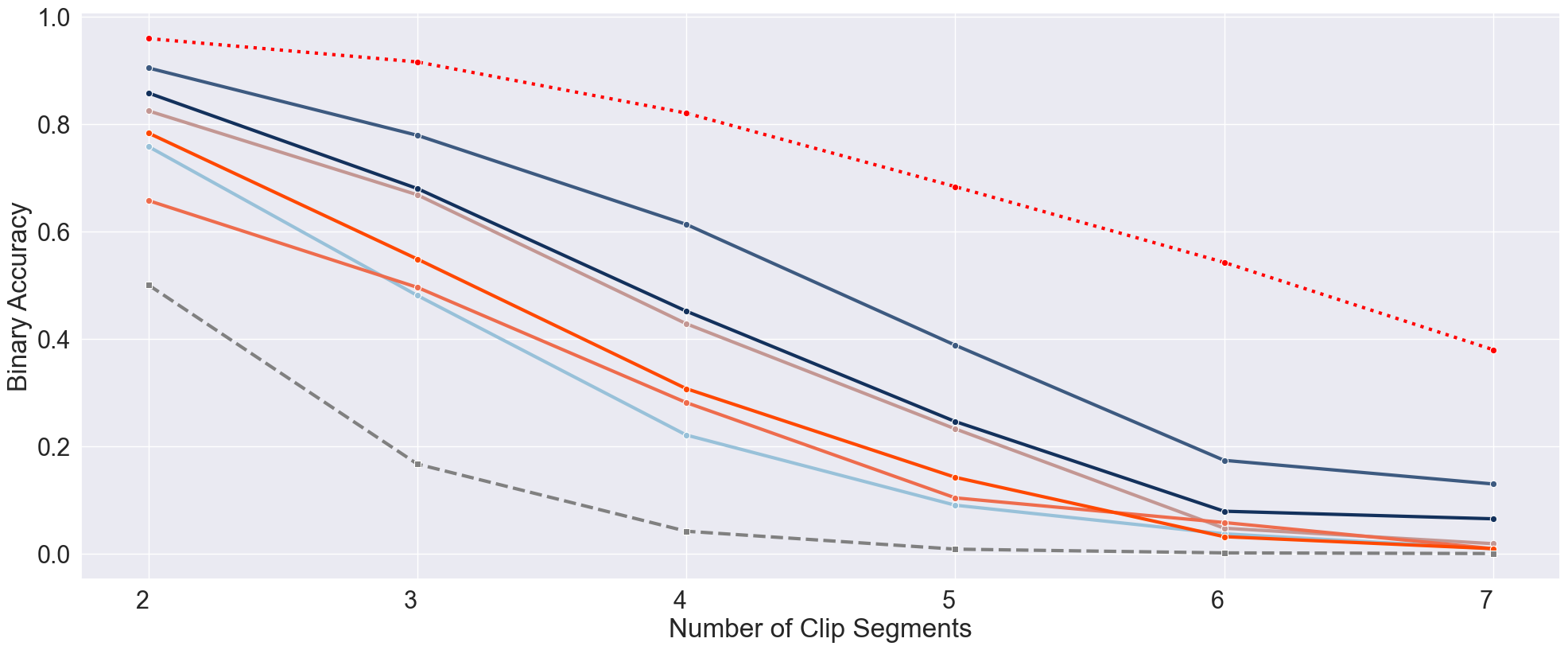}
            \caption{VLM performance by number of clips  with both video and text as input}
            \label{fig:vlm_performance_by_clip_length_labels}
        \end{subfigure}
        
        \caption{Binary accuracy performance of various state-of-the-art VLMs across different number of clips and modalities, compared to a human baseline (red dashed line) and a weighted random baseline (gray dashed line). Error bars represent the 95\% confidence interval (CI).}
        \label{fig:Appendix3}
    \end{center}
    \vskip -0.2in
\end{figure*}

\end{document}